%% file: zeroshotloss.tex
\documentclass[11pt,a4paper]{article}
\usepackage[colorlinks=true,linkcolor=blue]{hyperref}
\input{header.en.tex}

\usepackage{algorithm}
\usepackage{algorithmic}

\title{Tackling the Zero-Shot RL Loss Directly}
\author{Yann Ollivier}
\date{}

\newcommand{\betatest}{\beta_{\mathrm{test}}}
\newcommand{\losstest}{\ell_{\mathrm{test}}}
\newcommand{\Dir}{_\mathrm{Dir}}
\newcommand{\loss}{\mathcal{L}}
\newcommand{\orth}{_\mathrm{orth}}

\begin{document}

\maketitle

\begin{abstract}
Zero-shot reinforcement learning (RL) methods aim at instantly producing a behavior for an RL task in a given environment, from a description of the reward function.  These methods are usually tested by evaluating their average performance on a series of downstream tasks. Yet they cannot be trained directly for that objective, unless the distribution of downstream tasks is known.  Existing approaches either use other learning criteria \cite{borsa2018universal,zeroshot,allpolicies,visr}, or explicitly set a prior on downstream tasks, such as reward functions given by a random neural network \cite{frans2024unsupervised}.

Here we prove that the zero-shot RL loss can be optimized directly, for a range of non-informative priors such as white noise rewards, temporally smooth rewards, ``scattered'' sparse rewards, or a combination of those.

Thus, it is possible to learn the optimal zero-shot features algorithmically, for a wide mixture of priors.

Surprisingly, the white noise prior leads to an objective almost identical to the one in VISR \cite{visr}, via a different approach. This shows that some seemingly arbitrary choices in VISR, such as Von Mises--Fisher distributions, do maximize downstream performance. This also suggests more efficient ways to tackle the VISR objective.

Finally, we discuss some consequences and limitations of the zero-shot RL objective, such as its tendency to produce narrow optimal features if only using Gaussian dense reward priors.
\end{abstract}

\section{Introduction}

Zero-shot reinforcement learning (RL) methods aim at instantly producing
a behavior for an RL task in a given environment, from a description of
the reward function. This is done after an unsupervised training phase.
Such methods include, for instance, universal successor features (SFs,
\cite{borsa2018universal}) and the forward-backward framework (FB,
\cite{zeroshot, allpolicies}).

Zero-shot RL is usually tested by reporting average performance on
a series of downstream tasks: a reward function $r$ is sampled from a
distribution $\betatest$ of tasks, a reward representation $z=\Phi(r)$ is
computed, \footnote{A requirement of zero-shot RL is that this
computation should be scalable, with $z$ of reasonable size. Without a computational constraint, one
could just pre-compute all optimal policies of all possible downstream
tasks up to some degree of approximation.
} and a policy $\pi_z$ is applied, starting at some initial state
$s_0$. Thus, the reported performance is the expectation
\begin{equation}
\label{eq:testloss}
\E_{r\sim \betatest} \E_{s_0\sim \rho_0} V^{\pi_z}_r(s_0)
\end{equation}
where the value function $V^{\pi_z}_r(s_0)$ is the performance of policy
$\pi_z$ on the reward function $r$ when starting at $s_0$
(Section~\ref{sec:notation}).

Yet zero-shot RL methods are usually not trained by maximizing the
performance \eqref{eq:testloss}, because the distribution of downstream
tasks $\betatest$ is unknown. Other training criteria have to be
introduced, such as a finite-rank representation of long-term transition
probabilities in FB \cite{allpolicies}, or an information criterion on
policies $\pi_z$ in VISR \cite{visr}.

Alternatively, it is possible to explicitly set a prior
$\beta$ on downstream tasks, and optimize the criterion \eqref{eq:testloss} using that
prior instead of the true task distribution $\betatest$. 
This follows the machine learning
philosophy of ``follow the gradient of what you are actually doing'',
rather than made-up criteria.

For instance,
\cite{frans2024unsupervised} use random neural networks as a prior for
the downstream reward function $r$. This prior is parametric (it is
parameterized by the weights of a network), and it is unclear how
sensitive performance is to this choice.

Here:
\begin{itemize}
\item
We show that the zero-shot RL performance can be maximized directly
for a wide mixture of nonparametric, uninformative priors. This includes
dense reward priors such as white noise rewards, temporally smooth
rewards with a ``Dirichlet norm'' prior related to Laplacian
eigenfunctions, and sparse priors such as mixtures of a number of target
states with random weights.

Arguably, a mixture of such uninformative priors has the best chance of
covering the unknown test distribution $\betatest$. Note that learning
meaningful representations does not require informative priors on
downstream tasks:  environment dynamics lead to informative
representations even with non-informative priors.

This makes it possible to compute the best possible representations
for zero-shot RL by following the gradient of the criterion
\eqref{eq:testloss}. Notably, we can do this for dense reward priors
without
explicitly sampling a reward from the prior, which
would not be possible for infinite-dimensional priors such as white
noise.

\item We clarify the implicit priors on rewards in SFs: the SF strategy
implicitly relies on a \emph{white noise prior} on rewards
(Section~\ref{sec:SF}).

Doing so, we extend the SF framework to other priors, such as a prior
based on the Dirichlet norm, which introduces temporal smoothness related
to Laplacian eigenfunctions (Section~\ref{sec:priors}).

\item We show a surprising connection with VISR \cite{visr}: VISR
``almost'' computes the optimal zero-shot features for a white noise
prior (Section~\ref{sec:lossistractable}). 
(The ``almost'' comes from a minor change in the way the features are
normalized.)

This is unexpected, as VISR was not defined to maximize downstream
zero-shot performance. Instead, VISR was defined as a feedback loop
between a diversity method \cite{eysenbach2018diversity} and successor
features, by 
training a family of
policies $\pi_z$ that maximize the rewards $\transp{\phi}z$ for some
features $\phi$, and learning $\phi$ in turn by increasing $\transp{\phi}z$
at the places visited by $\pi_z$, thus creating specialization.

This newfound connection between VISR and downstream performance for a
white noise prior may justify some seemingly arbitrary choices in VISR,
such as its use of Von Mises--Fisher distributions.

The analysis also suggests more efficient ways to tackle the VISR
objective, notably, relying on occupation measures rather than Monte
Carlo sampling.

\item We derive further theoretical properties of the zero-shot RL loss.
Notably, the Bayesian viewpoint has no regularizing effect on the
policies learned: these policies are ``sharp'' in that they are
necessarily optimal policies for some particular task $r$
(Proposition~\ref{prop:posteriormean}).

This has some consequences for exploration in settings where the reward
is not exactly known: indeed, the zero-shot RL setting assumes that the
reward function is fully specified at test time (such as reaching a
particular goal or maximizing a particular quantity).

This can also produce unexpectedly narrow optimal features
(Section~\ref{sec:whatfeatures}) for some particular priors.

\item We discuss some limitations of the zero-shot RL setting, and
possible extensions.

\end{itemize}

%


\section{Setup, Notation, and Some Reward Priors}
\subsection{General Notation}
\label{sec:notation}

\paragraph{Markov decision process.}
We consider a reward-free Markov decision process (MDP) $\mathcal{M}=(S,A,P,\gamma)$
with state space $S$, action space $A$, transition probabilities
$P(s'|s,a)$ from state $s$ to $s'$ given action $a$,
and discount factor $0 < \gamma < 1$ \cite{sutton2018reinforcement}.
A policy $\pi$ is a function $\pi\from S\to \mathrm{Prob}(A)$ mapping a
state $s$ to the probabilities of actions in $A$.
Given
$(s_0,a_0)\in S\times A$ and a policy $\pi$,
we denote $\Pr(\cdot|s_0,a_0,\pi)$ and $\E[\cdot|s_0,a_0,\pi]$ the
probabilities and expectations under state-action sequences $(s_t,a_t)_{t
\geq 0}$ starting at $(s_0,a_0)$ and following policy $\pi$ in the
environment, defined by sampling $s_t\sim P( s_t|s_{t-1},a_{t-1})$ and
$a_t\sim \pi( a_t | s_t)$.
Given any reward
function $r\from S \to \R$, the $Q$-function of $\pi$ for $r$ is
$Q_r^\pi(s_0,a_0)\deq \sum_{t\geq 0} \gamma^t \E
[r(s_t)|s_0,a_0,\pi]$. The \emph{value function} of $\pi$ for $r$ is
$V_r^\pi(s)\deq \sum_{t\geq 0} \gamma^t \E
[r(s_t)|s_0,\pi]$.

We assume access to a dataset consisting of \emph{reward-free} observed
transitions $(s_t,a_t,s_{t+1})$ in the environment. We denote by $\rho$
the distribution of states $s_t$ in the training set.

\paragraph{Occupation measures.} We let $\rho_0$ be some distribution of initial states in the environment; if no such
distribution is available, we just take $\rho_0\deq \rho$.

\emph{Occupation measures} will pop up repeadtedly in our analysis. The
occupation measure $d_\pi$ of policy $\pi$ is a probability distribution over $S$, defined
for each $X\subset S$ as
\begin{equation}
\label{eq:dpi}
d_\pi(X)\deq (1-\gamma) \E_{s_0\sim \rho_0} \sum_{t\geq 0} \gamma^t \Pr(s_t\in
X|s_0,\pi).
\end{equation}
In particular, by construction,
\begin{equation}
\label{eq:Vsucc}
\E_{s\sim d_\pi} r(s)=(1-\gamma)
\E_{s_0\sim \rho_0} V^\pi_r(s_0).
\end{equation}

\subsection{The Zero-Shot RL Objective: Optimize Expected Downstream
Performance}

Existing zero-shot RL procedures proceed as follows: after an unsupervised,
reward-free pretraining phase, the agent is confronted with a reward $r$
(either via reward samples or via an explicit reward formula), computes a
task representation $z=\Phi(r)$ in a simple, fast way, then apply an
existing policy $\pi_z$. The map $\Phi$ from reward to task
representation, as well as the policies $\pi_z$, are learned during
pretraining.

Such methods are evaluated by running the policies $\pi_z$ on a number of
downstream tasks, and reporting the cumulated reward. Thus, if
$\betatest$ is the distribution of downstream tasks, the reported
loss is, in expectation,
\begin{equation}
\losstest(\Phi,\pi)= -\E_{r\sim \betatest} \E_{s_0\sim \rho_0}
V_r^{\pi_{\Phi(r)}}(s_0)
\end{equation}
where $\rho_0$ is the distribution of initial states used for testing.
This corresponds to
sampling a downstream task $r \sim \betatest$, computing $z=\Phi(r)$, and
running $\pi_z$ on reward $r$.

Usually the downstream task distribution $\betatest$ is unknown. Still, if we have a prior $\beta$ on rewards, a natural objective for
the pretraining phase is to minimize the loss
\begin{equation}
\label{eq:mainloss}
\ell_\beta(\Phi,\pi)\deq -\E_{r\sim \beta} \E_{s_0\sim \rho_0}
V_r^{\pi_{\Phi(r)}}(s_0)
\end{equation}
over $\Phi$ and $\pi$. The prior $\beta$ should ideally encompass the
unknown actual distribution $\betatest$ of downstream tasks.

Without computational constraints, this problem is theoretically ``easy'' to solve:
just precompute all optimal policies for all possible rewards. This
corresponds to $\Phi=\Id$, namely, a reward function $r$ is representated
by $z=r$ itself, and then $\pi_z$ should just be the optimal policy for
$r$. If the state space is continuous, $r$ and $z$ are
infinite-dimensional.

In practical methods, the task representation $z$ will be
finite-dimensional. This means some reward functions $r$ are necessarily
lumped together via $\Phi$, and determining the best way to do this
(e.g., for a fixed dimension of $z$)
becomes a nontrivial mathematical question. This is what we address in
the rest of the text.

\subsection{Some Uninformative Priors on Reward Functions}
\label{sec:priors}

We now introduce some priors on downstream tasks. Ideally, the prior
should cover the true distribution of tasks at test time.
Since this distribution is unknown, we try to consider the most
uninformative priors we could handle, in the hope this results in more
generic zero-shot performance.

We consider both dense and sparse reward priors. For dense rewards, we
include white noise, and a Gaussian process based on the Dirichlet norm,
which imposes more spatial smoothness on the rewards than white noise,
related to Laplacian eigenfunctions. For sparse rewards, we consider
random goal-reaching (reaching a target state specified at random), and
mixtures of several goals with random weights.

These are some of the most agnostic models we can find on an arbitrary
state equipped with an arbitrary probability distribution. All models are
built to have well-defined continuous-space limits, and still make sense
in an abstract state space equipped with a measure $\rho$. To avoid
excessive technicality, we restrict ourselves to the finite case in this
text.

Importantly, these priors rely on quantities that can be estimated from
the dataset (such as expectations under $\rho$). This is why we use
norms related to the dataset distribution $\rho$.

We will also use mixtures of these priors.

\subsubsection{Dense Reward Priors}

\paragraph{White noise prior.} This is defined as 
\begin{equation}
\beta(r)\propto
\exp(-\norm{r}^2_\rho/2)
\end{equation}
where $\norm{f}^2_\rho\deq \E_{s\sim \rho}
f(s)^2$.

This prior is very agnostic: the reward at every state is assumed to be
independent from every other state.

\paragraph{Dirichlet prior.} 
This is defined as 
\begin{equation}
\beta(r)\propto
\exp(-\norm{r}^2\Dir/2)
\end{equation}
where
\begin{equation}
\norm{f}\Dir^2\deq \E_{(s_t,a_t,s_{t+1})\sim \rho}\,
(f(s_t)-f(s_{t+1}))^2+\alpha \norm{f}^2_\rho
\end{equation}
where some $\alpha>0$ is used because the first term vanishes on constant $f$.

Contrary to white noise, this prior enforces some smoothness over
functions: the values at related states are closer.

The Dirichlet norm is directly related to Laplacian eigenfunctions.
Indeed, when $\rho$ is the invariant distribution of the policy in the dataset
\footnote{
More precisely, it is sufficient that the distributions of $s_t$ and $s_{t+1}$ under
the distribution $\rho$ in the dataset are the same. This does not
require the existence of a specific policy that produced the dataset. For
instance, if a dataset is a mixture of long trajectories from several
policies, then the laws of $s_t$ and of $s_{t+1}$ in the dataset will be
almost the same (up to neglecting the first and last state of each trajectory).}, one has
\begin{equation}
\norm{f}\Dir^2=2\langle f,\Delta f\rangle_\rho + \alpha \norm{f}^2_\rho
\end{equation}
where $\Delta \deq \Id - P_0$ is the Laplace operator of the
transition operator $P_0(s_{t+1}|s_t)$ of the policy implicitly defined
by the dataset.

\paragraph{General Gaussian priors.} To avoid proving the same results
separately for white noise and Dirichlet priors, we will more generally
use priors of the form
\begin{equation}
\beta(r)\propto \exp(-\norm{r}^2_K)
\end{equation}
where $\norm{f}^2_K$ denotes an arbitrary symmetric positive-definite
quadratic form on reward functions.

On a finite state space, this is equivalent
to $\exp(-\transp{r}Kr/2)$ for some p.s.d.\ matrix $K$ of size $\#S\times
\#S$. For instance, on
a finite state space, the white noise prior corresponds to
$K=\diag(\rho)$, and the Dirichlet prior is given by the matrix
$K=\E_{(s,s')\sim \rho}
[(\1_s-\1_{s'})\transp{(\1_s-\1_{s'})}]+\alpha \E_{s\sim \rho}[
\1_s\transp{\1_s}]$.

On infinite state spaces, this is an ``infinite-dimensional Gaussian''
whose formal definition involves having a consistent set of Gaussian
distributions in every finite-dimensional projection.

We will also use the associated dot
product $\langle{f,g}\rangle_K$. For instance,
\begin{equation}
\langle{f,g}\rangle\Dir=\E_{(s_t,a_t,s_{t+1})\sim
\rho}\,(f(s_t)-f(s_{t+1}))(g(s_t)-g(s_{t+1}))+\alpha f(s_t)g(s_t).
\end{equation}
Like $\norm{f}\Dir$, this can be estimated from the dataset.

\begin{rem}
In general, the optimal features are \emph{not} directly related to the
largest eigenvectors or singular vectors of $K$. For instance, the white
noise prior corresponds to $K=\diag(\rho)$, whose eigendecomposition is
independent of the dynamics of the environment, while optimal features
depend on the dynamics.
\end{rem}

\subsubsection{Sparse Reward Priors}
\label{sec:sparsepriors}

\paragraph{Random goal-reaching prior.} A \emph{goal-reaching} reward is
a reward that is nonzero only at a particular state, and $0$ everywhere
else.

If the prior $\beta$ on downstream tasks only includes goal-reaching
tasks (with some distribution of goals $g$), then arguably zero-shot RL
is not needed: it is better to just do goal-reaching, namely, directly
use $z=g$ as the task representation for goal $g$, and learn $Q(s,g)$ via
algorithms such as HER \cite{andrychowicz2017hindsight}.

But we want to mix goal-reaching with other priors, and find zero-shot RL
methods that can work in a mixture of different priors, hence the
interest of a general setup. So we formally define here a goal-reaching
prior.

In this model, we first
select a random state $s^\star\sim \rho$ in $S$. Then we put
a reward $1/\rho(s^\star)$ at $s^\star$, and $0$
everywhere else:
\begin{equation}
\label{eq:goalreaching}
r(s)=\frac{1}{\rho(s^\star)} \,\1_{s=s^\star}.
\end{equation}

The $1/\rho$ factor maintains $\int r\d \rho=1$. Without this scaling,
all $Q$-functions degenerate to $0$ in continuous spaces,
as discussed in
\cite{blier2021unbiased}. Indeed, if we omit this factor, and just set the reward to be $1$ at
a given goal state $s^\star\in S$ in a continuous space $S$, the probability of
exactly reaching that state with a stochastic policy is usually $0$, and all
$Q$-functions are $0$. Thanks to the $1/\rho$ factor, the
continuous limit is a \emph{Dirac function} reward, infinitely sparse,
corresponding to the limit of putting a reward $1$ in a small ball
$B(s^\star,\eps)$ of radius $\eps\to 0$ around $s^\star$, and rescaling by
$1/\rho(B(s^\star,\eps))$ to keep $\int r\d \rho=1$. This produces meaningful,
nonzero $Q$-functions in the continuous limit \cite{blier2021unbiased}.

This model combines well with successor features or the FB framework:
indeed, this model satisfies
\begin{equation}
\E_{s\sim \rho} [r(s)\phi(s)]=\phi(s^\star)
\end{equation}
(both in finite spaces and in the continuous-space limit). This is useful
in conjunction with the SF formulas such as \eqref{eq:SF} in Section~\ref{sec:SF}.

\paragraph{Scattered random reward prior.} We extend the random
goal-reaching prior to rewards comprising several goals with various
weights, where the weights may be random and may be positive or negative.

Generally speaking, we will call \emph{scattered random reward prior} any
prior which consists in first choosing an integer $k\geq 0$ according to
some probability distribution, then choosing $k$ goal states
$(s^\star_i)_{1\leq i \leq k}\sim \rho$ and $k$ random weights
$w_i,\ldots,w_k$ according to some fixed probability distribution on
$\R$, and setting
\begin{equation}
r(s)=c_k \sum_{i=1}^k 
\frac{w_i}{\rho(s^\star_i)} \,\1_{s=s^\star_i}
\end{equation}
namely, a sum of $k$ goal-reaching rewards \eqref{eq:goalreaching}.

A suitable scaling factor $c_k$ can sometimes produce more
meaningful behavior for large $k$. For instance, if we take $w_i\sim N(0,1)$ and
$c_k=1/\sqrt{k}$, and let $k\to \infty$, then this prior tends to the white noise
prior above.

Therefore, scattered random reward priors can be seen as interpolating
between the pure goal-reaching and white noise priors.

\section{Algorithmic Tractability of the Zero-Shot RL Loss}

\subsection{The Optimal Policies Given a Representation $\Phi$}

Here we work out half of the objective \eqref{eq:mainloss}: what are the
optimal policies
$\pi_z$ if the task representation $\Phi$ is
known?

\begin{prop}[ (Policies must be optimal for the mean posterior reward
knowing $z$)]
\label{prop:posteriormean}
For each $z$, define
\begin{equation}
\label{eq:postreward}
r_z\deq \E_{r|\Phi(r)=z} [r]
\end{equation}
the mean reward function knowing $\Phi(r)=z$ under the prior $\beta$. Let
also $\beta_z$ be the distribution of $z=\Phi(r)$ when sampling $r\sim
\beta$.

Then
\begin{equation}
\ell_\beta(\Phi,\pi)=-\E_{z\sim \beta_z}\,\E_{s_0\sim \rho_0} V^{\pi_z}_{r_z}(s_0).
\end{equation}

Consequently, given the representation $\Phi$, for every $z$, the best policy $\pi_z$ is the optimal
policy $\pi^\star_{r_z}$ for reward $r_z$.
\end{prop}

So, in this model, the optimal zero-shot policies have no induced stochasticity to account for
uncertainties. This holds even if there is noise in the computation of
$z$. The full point of zero-shot RL is to decide which rewards to lump together under the
same policy.

This does not hold if one includes variance over $r$ in the main loss
\eqref{eq:mainloss} (Section~\ref{sec:varianceregul}).

The value of $r_z$ can be derived explicitly for some priors (Gaussian
priors and linear $\Phi$), which we now turn to. Other priors
(goal-oriented or scattered
random rewards) require a slightly different approach
(Section~\ref{sec:phisparse}).

\subsection{Linear Task Representations $\Phi$}
\label{sec:SF}

We now emphasize the case of \emph{linear} task representations
$\Phi$, because it corresponds to successor features and to the
forward-backward framework, which are the most successful zero-shot RL
approaches to date. (See Section~\ref{sec:future} for nonlinear $\Phi$.)

The easiest-to-compute
task representations $z=\Phi(r)$ are linear functions of $r$. Any such
finite-dimensional function $\Phi$ is given by integrating
the reward against some features $\phi=(\phi_{i}(s))_{i=1,\ldots,k}$:
\begin{equation}
z=(\Cov \phi)^{-1} \E_{s\sim \rho} \,r(s)\phi(s)
\end{equation}
where we include a preconditioning by $(\Cov \phi)^{-1}$ as in SFs. \footnote{
Including $(\Cov \phi)^{-1}$ from the start (as opposed to
$z=\E_{s\sim \rho} \,r(s)\phi(s)$ as in FB) is more adapted to
distribution shifts. Indeed, for rewards in the span of
$\phi$, then the reward representation 
$z$ is independent of the distribution $\rho$ of states used for the
computation.
} 
Here all covariances are expressed with respect to the state
distribution $\rho$: $\Cov \phi\deq \E_{s\sim \rho} \,\phi(s)\transp{\phi(s)}$.

By Proposition~\ref{prop:posteriormean}, given the features $\phi$, the
best policies are the optimal policies for the rewards $r_z$. So we 
have to compute $r_z$. Then the policies can be learned, e.g.,\ via
$Q$-learning for each $z$.

The following result specifies the value of $r_z$ and hence the optimal
policies given the features, but does not yet say how to choose the
features $\phi$: this is covered in the next sections.

\begin{prop}[ (Linear task representations and white noise prior)]
\label{prop:linearpostmean}
Assume the reward representation $z=\Phi(r)$ is given by the successor
feature model
\begin{equation}
\label{eq:SF}
z=(\Cov \phi)^{-1} \E_{s\sim \rho} \,r(s)\phi(s)
\end{equation}
using some linearly independent features $\phi\from S \to \R^d$.

Then, for the white noise prior on rewards, the posterior mean 
reward $r_z$ \eqref{eq:postreward} is
\begin{equation}
\label{eq:rz}
r_z(s)=\transp{z}\phi(s).
\end{equation}
Therefore, by Proposition~\ref{prop:posteriormean}, for a given $\phi$, the policies $\pi_z$ that optimize the
zero-shot RL loss \eqref{eq:mainloss} are the
optimal policies for reward $\transp{z}\phi$, for each $z$.

Moreover, under these assumptions, the distribution $\beta_z$ of $z$ is a centered Gaussian with
covariance matrix $(\Cov \phi)^{-1}$.
\end{prop}

This proposition gives a justification for part of the strategy behind
successor features, namely, projecting the reward onto the features and
applying the optimal policy for the projected reward. This is optimal on
average under
an implicit \emph{white noise prior} on rewards. 


In the forward-backward framework (FB), the task representation $z$ is
computed as $z=\E_{s\sim \rho} r(s)B(s)$ with features $B$. This is the
same as \eqref{eq:SF} up to the change of variables by $(\Cov
\phi)^{-1}$. Therefore, this result strongly suggests to train policies
$\pi_z$ for the rewards $\transp{z}(\Cov B)^{-1}B$. This contrasts with
the FB framework, in which the policies $\pi_z$ are defined through the
forward function $F$. The two coincide only if the training of $F$ is
perfect. \footnote{because in that case, $F$ contains the successor
features of $(\Cov B)^{-1}B$, by one of the results in
\cite{allpolicies}.}

In general, the proposition is \emph{not} true for other reward priors, such as
random goal-reaching. \footnote{For instance, take any set of features
such that $\phi\from S\to \R^d$ is injective, such as $\phi=\Id$. Take
for $\beta$ the goal-reaching prior. Then for reaching a goal $g$, the
reward $r$ is a Dirac at $g$ so that
$\E [r\phi]=\phi(g)$ and
$z=C^{-1}g$ with $C$ the covariance matrix. Since the map $g\to z$ is
bijective, it conveys full knowledge of the task for this prior, and the
posterior mean $r_z$ is just the single reward for reaching $g$.}
Still, \eqref{eq:rz} also
holds for any Gaussian prior on rewards such that the
components of $r$ along $\phi$ and its $L^2(\rho)$-orthogonal are independent, namely,
$r(s)=\transp{\theta_1}\phi(s)+\transp{\theta_2}\xi(s)$ where $\xi$ are
any features such that $\E_{s\sim \rho}[\phi(s)\transp{\xi(s)}]=0$ and
$\theta_1$, $\theta_2$ are independent Gaussian vectors with any
covariance matrix. But this cannot be used to optimize the features
$\phi$, because this condition depends on $\phi$ itself so it does not
represent a fixed prior for the loss \eqref{eq:mainloss}.

\bigskip

This result extends to the more general case of arbitrary Gaussian priors
given by a metric $\norm{\cdot}_K$: we just have to compute $z$ by a formula
involving this norm, instead of the SF formula
\eqref{eq:SF}. 

This is especially relevant if $K$
can be computed from expectations over the dataset, as with the Dirichlet
prior: this results in an SF-like approach, but relying on a different
implicit prior instead of the white noise prior on rewards.

\begin{prop}[ (Linear representations with arbitrary Gaussian prior)]
\label{prop:gaussianprior}
Assume that the prior on rewards is
\begin{equation}
\beta(r)\propto \exp(-\tfrac12 \norm{r}^2_K)
\end{equation}
for some Euclidean norm $\norm{\cdot}_K$ on the space of rewards.

Assume the reward representation $z=\Phi(r)$ is computed as
\begin{equation}
\label{eq:genSF}
z=C^{-1}  \langle r,\phi\rangle_K
\end{equation}
using some linearly independent features $\phi\from S \to \R^d$,
where $C$ is the $k\times k$ matrix with entries $C_{ij}=\langle
\phi_i,\phi_j\rangle_K$. Namely, $z$ contains the weights of
the $L^2(\norm{\cdot}_K)$-orthogonal projection of $r$ onto the features
$\phi$.

Then the posterior mean 
reward $r_z$ given $z$ is
\begin{equation}
\label{eq:rzD}
r_z(s)=\transp{z}\phi(s)
\end{equation}
Therefore, by Proposition~\ref{prop:posteriormean}, for a given $\phi$,
the policies $\pi_z$ that optimize the zero-shot RL loss
\eqref{eq:mainloss} are the optimal policies for reward $\transp{z}\phi$,
for each $z$.

Moreover, the distribution $\beta_z$ of $z$ is Gaussian with
covariance matrix $(\langle \phi,\phi\rangle_K)^{-1}$.
\end{prop}

For instance, with the Dirichlet prior, we have
\begin{equation}
\langle\phi ,r\rangle\Dir = \E_{(s_t,s_{t+1})\sim \rho}\,
(r(s_t)-r(s_{t+1}))(\phi(s_t)-\phi(s_{t+1}))+ \alpha \E_{s\sim\rho}\,
r(s)\phi(s)
\end{equation}
and
\begin{equation}
\langle \phi , \phi\rangle\Dir= \E_{(s_t,s_{t+1})\sim
\rho}\,(\phi(s_t)-\phi(s_{t+1}))\transp{(\phi(s_t)-\phi(s_{t+1}))}
+\alpha\E_{s\sim \rho}\, \phi(s)\transp{\phi(s)}
\end{equation}
and so $z$ can be estimated from
samples. This gives rise to a Dirichlet-prior-based version of successor
features. \footnote{Depending on how the reward is specified for zero-shot
RL, in some situations, we might not have access to both
$r(s_t)$ and $r(s_{t+1})$. And for goal-oriented tasks, we usually don't have access
to $\phi(s_{t+1})$, the state visited one step after reaching the goal.

This contrasts with basic successor features, for which setting a goal
state
$s^\star$ just gives $z\propto (\Cov \phi)^{-1} \phi(s^\star)$.
}

\begin{rem}
It
is also possible to train a Gaussian $\exp(-\norm{r}^2_K)$ prior while
using features $z=(\E \phi\transp\phi)^{-1} \E r\phi$ that do not use the $K$-norm.
But in that case the expression for the posterior mean $r_z$
is much more complex
and requires inverting $K$. 
\end{rem}

\subsection{The Zero-Shot Loss is Tractable for Linear Representations}
\label{sec:lossistractable}

These results for fixed $\phi$ pave the way to
computing the gradient of the zero-shot loss
\eqref{eq:mainloss} with respect to $\phi$: putting together all the
ingredients yields the following result.

\begin{thm}[ (Zero-shot RL loss for linear task representations)]
\label{thm:main}
Assume that the prior $\beta$ on reward functions $r$ is $\beta(r)\propto
\exp(-\tfrac12 \norm{r}^2_K)$ for some Euclidean norm $\norm{\cdot}_K$.
Assume that the reward representation $z=\Phi(r)$ is computed as in successor
features \eqref{eq:genSF} using the norm $\norm{\cdot}_K$, namely,
\begin{equation}
z=C^{-1} \langle r,\phi\rangle_K
\end{equation}
where $\phi\from S\to \R^d$ are linearly independent features, and where
$C$ is the matrix with entries $C_{ij}= \langle \phi_i,\phi_j\rangle_K$.

Then the zero-shot RL loss \eqref{eq:mainloss} is
\begin{equation}
\label{eq:tractableloss}
\ell_\beta(\Phi,\pi)=-\frac1{1-\gamma} \E_{z\sim N(0,C^{-1})} \,\E_{s\sim
d_{\pi_z}}\transp{\phi(s)}z
\end{equation}
where $d_{\pi_z}$ is the occupation measure \eqref{eq:dpi} of policy
$\pi_z$.

Moreover, the optimal $\pi_z$ given $\Phi$ is the optimal policy for
reward $r_z(s)\deq \transp{\phi(s)}z$.
\end{thm}


\paragraph{Relationship with VISR \cite{visr}: VISR almost optimizes
expected downstream performance under a white noise prior.} Surprisingly, the loss
\eqref{eq:tractableloss} is very close to the loss optimized in VISR,
although VISR was built in a different way with no formal connection to
expected downstream task performance.

VISR is a criterion to build features $\phi$ for successor features. It
works with a set of features $\phi$ and policies $\pi_z$. Each policy
$\pi_z$ is the optimal policy for reward function $\transp{\phi}z$. The
features $\phi$ are chosen to maximize the mutual information between $z$
and the states $s$ visited by $\pi_z$; more exactly, the states $s$ are
assumed to be observed only through $\phi(s)$, and the distribution of
$z$ knowing $\phi(s)$ is assumed to follow a Von Mises--Fisher
distribution $\exp(\transp{\phi}z)$ (this is chosen for convenience so
that the log-likelihood $\transp{\phi}z$ matches with the reward). The
features $\phi$ attempt to maximize the mutual information under this
model of $z$ given $s$; this mutual information is estimated via a
variational lower bound. We refer to the VISR paper \cite{visr} for
further details.

Yet it turns out Algorithm 1 in \cite{visr} optimizes the loss
\eqref{eq:tractableloss} above, except for a difference in the way
$z$ and $\phi$ are normalized. More precisely, the VISR algorithm
consists in:
\begin{enumerate}
\item Sampling a hidden vector $z$ (denoted $w$ in \cite{visr}).
\item Training the policy $\pi_z$ to optimize the reward
$\transp{\phi}z$. This is done in VISR via the computation of the successor
features $\psi$ of $\phi$.
\item Running the policy $\pi_z$ to get a sequence of states $s_t$, whose
distribution is thus $d_{\pi_z}$.
\item Updating the features $\phi$ to minimize $-\transp{\phi(s_t)}z$.
\end{enumerate}

VISR ``almost'' optimizes the loss \eqref{eq:tractableloss}:
the only difference between VISR and Theorem~\ref{thm:main} lies in the
normalization of $z$ and $\phi$. In Theorem~\ref{thm:main}, we sample $z$ from $N(0,C^{-1})$
where $C$ is the covariance matrix of $\phi$, and we have no constraint
on $\phi$. In VISR, $z$ is sampled from $N(0,\Id)$ then normalized to
unit length, and the features $\phi$ use a normalized output layer so
that $\norm{\phi(s)}=1$ for any state $s$.

Normalization is necessary in VISR: otherwise, the loss of $\phi$ can be
brought to $0$ by downscaling $\phi$. On the other hand, in
Theorem~\ref{thm:main}, if we downscale $\phi$, the distribution $z\sim
N(0,C^{-1})$ gets \emph{up}scaled by the same factor so $\transp{\phi}z$
is unchanged. This emphasizes the role of sampling $z$ with covariance
matrix
$C^{-1}$. Also note that the normalization $\norm{\phi(s)}=1$ in VISR
does \emph{not} imply that the covariance matrix of $\phi$ is $C=\Id$. So there is
a slight mismatch between the VISR objective and the zero-shot RL loss. 

Still, Theorem~\ref{thm:main} proves that \emph{VISR ``almost'' optimizes the
expected downstream performance of $\pi_z$ under a white noise prior on
reward functions}, where the ``almost'' accounts for the difference in
normalization and covariance of $z$. This is surprising, as expected downstream
performance was not explicitly used to derive VISR.

\subsection{Algorithms for Optimizing the Representation $\phi$}
\label{sec:phiopt}

A generic VISR-like algorithm to optimize the zero-shot RL loss
\eqref{eq:tractableloss} in Theorem~\ref{thm:main} may have the following
structure:
\begin{enumerate}
\item Sample a minibatch of $z$ values.
\item Do a policy optimization step to bring $\pi_z$ closer to the
optimal policy for reward $\transp{\phi(s)}z$.
\item Estimate the occupation measures $d_{\pi_z}$ of $\pi_z$.
\item Do a gradient step on $\phi$ using the loss
\eqref{eq:tractableloss}.
\item Iterate.
\end{enumerate}

We present one possible such algorithm in Algorithm~\ref{algo:main}. It
departs from VISR in three ways:
\begin{itemize}
\item Fixing normalization and influence of $C$: sampling  $z$ from
$N(0,C^{-1})$. An extra complication occurs: since $C$ depends on $\phi$,
it is necessary
to estimate the gradients coming from $C^{-1}=(\langle
\phi,\phi\rangle_K)^{-1}$ when taking gradients with respect to $\phi$.

This ensures we exactly optimize te zero-shot RL loss
\eqref{eq:tractableloss}.

\item Estimating a model of the occupation measures $d_{\pi_z}$.  VISR
obtains sample states $s\sim d_{\pi_z}$ by running trajectories of
$\pi_z$ and using a Monte Carlo estimate by averaging over these
trajectories. This both suffers from high variance and limits
applicability to the online RL setup, since interactions with the
environment are needed during training.

Instead, learning a model of $d_{\pi_z}$ allows Algorithm~\ref{algo:main}
to run in an offline RL setting. It should also result in larger bias but
smaller variance with respect to Monte Carlo sampling from $d_{\pi_z}$.

\item Simplifying the learning of $\pi_z$: this can be done using any
$Q$-learning algorithm with $z$-dependent $Q$-function $Q(s,a,z)$ for 
reward $\transp{\phi(s)}z$. It does not have to use the successor
features of $\phi$ as in VISR.
\end{itemize}

\newcommand{\algindent}{\STATE\hspace{\algorithmicindent}}

  \begin{algorithm}[tb]
    \small
    \caption{One possible algorithm to optimize the zero-shot RL loss
    \eqref{eq:tractableloss}}
    \label{algo:main}
 \begin{algorithmic}
    \STATE \textbf{Input:}
    \\Dataset of transitions $(s_t,a_t,s_{t+1})$
    with distribution $\rho$.
    \\Norm $\norm{\cdot}_K$ on features (default:
    $\norm{\phi}_K^2\deq \E_{s\sim \rho} \abs{\phi(s)}^2$), and
    associated dot product.
    \\Weights $\lambda_C\in \{0,1\}, \lambda\orth\geq0$ for auxiliary
    losses.
    \\Online EMA weights $\beta_t\in (0,1)$ to estimate $C$.
    \STATE \textbf{Output:}\\
    Trained features $\phi_1,\ldots,\phi_d$ with their covariance
    matrix $C$.
    \\Trained policies $\pi_z$.
    \WHILE{not done}
    \STATE Update covariance matrix $C$ via EMA:
    $C_{ij}\gets \beta_t
    C_{ij}+(1-\beta_t)\langle \phi_i,\phi_j\rangle_K$
    \STATE Sample a minibatch of values of $z$: $z\sim N(0,C^{-1})$
    \STATE Update a $Q$-function $Q(s,a,z)$ and policy $\pi_z(a|s)$ for
    reward $\transp{\phi(s)}z$, using any RL algorithm
    \STATE Update the occupation measure model $d(s,z)$ via one step of
    Algorithm~\ref{algo:occupation}
    \STATE Sample a minibatch of states $s$ from the dataset, and
    update $\phi$ with the loss
    \algindent $\loss(\phi)=-d(s,z)\transp{\phi(s)}z+\lambda_C\,
    \loss_C(\phi,s,z)+\lambda\orth\,\loss\orth(\phi)$
    \\where $\loss_C$ and $\loss\orth$ are the auxiliary losses
    \eqref{eq:lossC} and \eqref{eq:lossorth} respectively
    \ENDWHILE
    \STATE \textbf{Deployment:}
    \\Once the reward function $r$ is known:
    \STATE Estimate $\langle r,\phi_1\rangle_K$, \ldots $\langle
    r,\phi_d\rangle_K$
    \STATE Set $z=C^{-1}\langle r,\phi\rangle_K$
    \STATE Apply policy $\pi_z$
 \end{algorithmic}
 \end{algorithm}

Let us further discuss two of these points (gradients coming from $C$, and
estimating $d_{\pi_z}$). The exact derivations are included in
Appendix~\ref{sec:proofs}.

\paragraph{Learning the occupation measures $d_{\pi_z}$.}
Instead of explicitly running the policy $\pi_z$ as in VISR, a number of techniques allow for direct estimation of the
density of $d_{\pi_z}$.

Indeed, $d_{\pi_z}(s)$ is the average over $s_0\sim \rho_0$ of the
\emph{successor measures} $M^{\pi_z}(s_0,a_0,s)$, multiplied by
$(1-\gamma)$. We refer to \cite{successorstates} or to
Appendix~\ref{sec:proofs}
for successor measures: intuitively, $M^{\pi_z}(s_0,a_0,s)$ encodes the
expected amount of time spent at $s$ if starting at $(s_0,a_0)$ and
running $\pi_z$.

Algorithm~\ref{algo:occupation} first learns a model $m(s_0,a_0,s,z)$ of
the successor measure, using one of the methods from
\cite{successorstates} (the measure-valued Bellman equation satisfied by
successor measures). Then it averages the result over $s_0$ and $a_0$ to
obtained the model $d(s,z)$ of the occupation measure $d_{\pi_z}$. The
mathematical derivations are given in Appendix~\ref{sec:proofs}. The
model $m(s_0,a_0,s,z)$ may take any form; a particular case is a
finite-rank approximation $m(s_0,a_0,s,z)=\transp{F(s_0,a_0,z)}B(s,z)$
similar to the forward-backward representation from \cite{allpolicies},
except that here $B$ can be $z$-dependent.
\footnote{
A model $m(s_0,a_0,s,z)=\transp{F(s_0,a_0,z)}B(s)$, with $B$ independent
of $z$, would be too
restrictive here: in this model, everything is projected onto the span of
$B$, and the optimal $\phi$ is just $B$.
}

  \begin{algorithm}[tb]
    \small
    \caption{One possible algorithm to estimate occupation measures
    $d(s,z)$}
    \label{algo:occupation}
 \begin{algorithmic}
    \STATE \textbf{Input:} Dataset of transitions $(s_t,a_t,s_{t+1})$
    with distribution $\rho$.
    \STATE Distribution of initial states $\rho_0$
    (default: $\rho_0=\rho$).
    \STATE Policies $\pi_z(a|s)$.
    \STATE Covariance matrix $C$ for sampling $z$.
    \STATE \textbf{Output:} Trained occupation model $d(s,z)$.
    \WHILE{not done}
    \STATE Sample a minibatch of values of $z$: $z\sim N(0,C^{-1})$
    \STATE Sample a minibatch of transitions $(s_t,a_t,s_{t+1})\sim \rho$
    \STATE Sample actions $a_{t+1}\sim \pi_z(a_{t+1}|s_{t+1})$
    \STATE Sample a minibatch of states $s'\sim \rho$
    \STATE Update the successor measure model $m(s_t,a_t,s',z)$ with the loss
    \algindent $\loss(m)=
\left(m(s_t,a_t,s',z)-\gamma
\bar m(s_{t+1},a_{t+1},s',z)\right)^2-2m(s_t,a_t,s_t,z)$ with $\bar m$ a
target network version of $m$ (using EMA of parameters of $m$ and a
stop-grad)
    \STATE Sample a minibatch of initial states $s_0\sim \rho_0$ and
    actions $a_0\sim \pi_z(a_0|s_0)$
    \STATE Update the occupation measure model $d(s,z)$ with the loss
    \algindent $\loss(d)=\left(d(s',z)-(1-\gamma)m(s_0,a_0,s',z)\right)^2$
    \ENDWHILE
 \end{algorithmic}
 \end{algorithm}

\paragraph{Dealing with the covariance matrix $C$.} In the loss
\eqref{eq:tractableloss}, the variable $z$ is sampled from $z\sim
N(0,C^{-1})$. Since $C$ depends on $\phi$, this produces extra terms when
attempting to optimize the loss over $\phi$.

Here a reparameterization trick $z\gets C^{1/2}z$ is inconvenient,
because it still requires computing the gradient of $C^{-1/2}$ with
respect to $C$, and this requires inverting a $d^2\times d^2$ matrix, not
just a $d\times d$ matrix.

Instead, two other strategies are possible:
\begin{enumerate}
\item Only work with orthonormal features, i.e., impose $C=\Id$ at all
times. This is possible without loss of generality, because zero-shot RL
with linear features only depends on the linear span of the features.

In practice, this can be done by imposing a Lagrange multiplier for the
constraint $C=\Id$. This means adding a loss term $\lambda\orth\,
\loss\orth(\phi)$ in the algorithm, where $\lambda\orth$ is a large
weight, and where
\begin{align}
\label{eq:lossorth}
\loss\orth(\phi)& \deq  \norm{\langle
\phi,\phi\rangle_K-\Id}^2_{\mathrm{Frobenius}}
\\&=-2\sum_i \norm{\phi_i}_K^2
+\sum_{ij} \left(\langle \phi_i,\phi_j\rangle_K\right)^2
+\mathrm{cst}
\end{align}
is the loss associated with violating the constraint $C=\Id$.
This option corresponds to $\lambda_C=0$ in Algorithm~\ref{algo:main}.

With $\norm{\cdot}_K=\norm{\cdot}_\rho$, this loss simplifies to
\begin{equation}
\loss\orth(\phi)=\E_{s\sim \rho,\,s'\sim \rho}\left[
\left(\transp{\phi(s)}\phi(s')\right)^2-\norm{\phi(s)}^2-\norm{\phi(s')}^2
\right]+\mathrm{cst}
\end{equation}
similarly to the orthonormalization loss for $B$ in \cite{zeroshot}.

Even with a large weight $\lambda\orth$, the condition $C=\Id$ will be
satisfied only approximately. Thus we still include $C$ in the algorithm.

When using the orthonormalization loss $\loss\orth$ with a large weight,
it is better if $\phi$ is initialized so that $C$ is not too far from
$\Id$.

\item The second option provides an exact estimation of the gradient of
$C$. 
This is carried out in Appendix~\ref{sec:proofs}, and results in the
following loss $\loss_C$ included in Algorithm~\ref{algo:main}:
\begin{equation}
\label{eq:lossC}
\loss_C(\phi,s,z)\deq
\tfrac12
d(s,z)\left(\transp{\bar \phi(s)}z\right)
\sum_{ij} \left((\bar C^{-1})_{ij}-z_iz_j\right)\langle
\phi_i,\phi_j\rangle_K
\end{equation}
where $\bar \phi$ and $\bar C$ are stop-grad versions of $\phi$ and $C$,
respectively.

If $\norm{\cdot}_K=\norm{\cdot}_\rho$, this can be estimated
as
\begin{equation}
\loss_C(\phi,s,z)=\tfrac12 d(s,z)\left(\transp{\bar \phi(s)}z\right)
\,\E_{s'\sim \rho} \left[\transp{\phi(s')}\bar
C^{-1}\phi(s')-(\transp{\phi(s')}z)^2
\right]
\end{equation}

Even if using the loss $\loss_C$, we still recommend to include a loss
$\loss\orth$, for numerical reasons to keep $\phi$ within a reasonable
numerical range. \footnote{When $\loss_C$ is included, mathematically $\loss\orth$
has no effect since everything only depends on the span of $\phi$ and not
$\phi$ itself. But numerically it will be more convenient to keep $\phi$
well-conditioned.}
\end{enumerate}

These results make it possible to optimize the features $\phi$ for a
Gaussian prior on downstream tasks. We now turn to other priors.

\subsection{Learning the Optimal Features for Sparse Reward Priors}
\label{sec:phisparse}

We now turn to the sparse reward priors from
Section~\ref{sec:sparsepriors}. Since goal-reaching is a special case of
scattered random rewards (with $k=1$), we only deal with the latter.
Namely, we consider sparse rewards of the form
\begin{equation}
\label{eq:sparsereward}
r=c_k \sum_{i=1}^k w_i \,\delta_{s^\star_i}
\end{equation}
where $\delta_{s^\star}(s)\deq
\1_{s=s^\star}/\rho(s^\star)$
is the Dirac sparse reward \footnote{Dirac with respect to the measure $\rho$,
namely, $\E_\rho [f.\delta_{s^\star}]=f(s^\star)$. In
particular, $\E_\rho \delta_{s^\star}=1$.} at
$s^\star$ as defined in Section~\ref{sec:sparsepriors}, 
$k$ is an integer following some probability distribution,
$(s^\star_i)_{1\leq i \leq k}$ are goal states sampled
from the data distribution $\rho$, the $w_i$ are weights sampled from
some distribution on $\R$, and $c_k$ is a scaling factor. A typical example is $w_i\sim N(0,1)$ and
$c_k=1/\sqrt{k}$.

Arguably, with such a model, we could just send the full reward
description $(s^\star_i,w_i)_{1\leq i \leq k}$ to a $Q$-function
or policy model. However, our goal is
to be able to mix several types of priors on rewards
(Section~\ref{sec:mixing}): we want to find zero-shot RL methods that
work both for dense and sparse rewards. Therefore, we describe a method
whose structure is closer to that of the previous sections, by learning
optimal features $\phi$.

\begin{prop}
\label{prop:sparseloss}
Let $\beta$ be a prior on sparse rewards of the type
\eqref{eq:sparsereward}, for some distribution of $(k,(w_i),c_k)$ and
where each $(s^\star_i,a^\star_i)$ has distribution $\rho$.

Assume that the reward representation $z=\Phi(r)$ is computed as in successor
features \eqref{eq:genSF} using the norm $\norm{\cdot}_K$, namely,
\begin{equation}
z=C(\phi)^{-1} \langle r,\phi\rangle_K
\end{equation}
where $\phi\from S\to \R^d$ are linearly independent features, and where
$C(\phi)$ is the matrix with entries $C_{ij}= \langle \phi_i,\phi_j\rangle_K$.

Then the zero-shot RL loss $\ell_\beta(\Phi,\pi)$ satisfies
\begin{equation}
\ell_\beta(\Phi,\pi)=-\frac{1}{1-\gamma} \E_{k,\, s^\star_i,\,w_i}\,
\sum_{i=1}^k
c_k w_i\,
d\left(s^\ast_i,z(\phi)\right)
\end{equation}
where
\begin{equation}
z(\phi)=\sum_j c_kw_j C(\phi)^{-1} \langle \delta_{s^\star_j},\phi\rangle_K
\end{equation}
and where $d(s,z)$ is the density of $d_{\pi_z}(s)$ with respect to the data
distribution $\rho$.
\end{prop}

The density $d(s,z)$ is the same as in Algorithm~\ref{algo:main}, and can
be learned via Algorithm~\ref{algo:occupation}.

Algorithm~\ref{algo:sparse} instantiates this result for the case where
$\norm{\cdot}_K=\norm{\cdot}_\rho$. In that case, we have
\begin{equation}
\langle \delta_{s^\star_j},\phi\rangle_\rho=\phi(s^\star_j)
\end{equation}
which simplifies the expression for $z$.

Two points in Algorithm~\ref{algo:sparse} are tricky. The first is how to compute the
gradient of $z(\phi)$ with respect to $\phi$, and in particular the
gradient of $C(\phi)^{-1}$. In Algorithm~\ref{algo:sparse}, we have used
that $C=\E_{s'} \phi(s')\transp{\phi(s')}$ when
$\norm{\cdot}_K=\norm{\cdot}_\rho$. We have included an extra term
\begin{equation}
\bar C^{-1}
    \left(\bar\phi(s')\transp{\bar\phi(s')}-\phi(s')\transp{\phi(s')}\right)\bar
    C^{-1}\bar \phi(s^\star_i)
\end{equation}
which evaluates to $0$ in the forward pass (since $\bar\phi=\phi$) but
provides the correct gradients with respect to $C(\phi)^{-1}$ in the backward pass.

The second tricky point is how to update the $Q$-function for the sparse
reward. Here we have directly applied the results from
\cite{blier2021unbiased} for $Q$-learning with Dirac rewards such as
\eqref{eq:sparsereward}. This point
is important when mixing different priors (Section~\ref{sec:mixing}): the
$Q$-functions for different priors should be updated in a consistent way,
(e.g., all updated using the Bellman loss for their respective rewards).

  \begin{algorithm}[h!]
    \small
    \caption{One possible algorithm to optimize the zero-shot RL loss
    with sparse rewards \eqref{eq:sparsereward}}
    \label{algo:sparse}
 \begin{algorithmic}
    \STATE \textbf{Input:}
    \\Dataset of transitions $(s_t,a_t,s_{t+1})$
    with distribution $\rho$.
    \\Online EMA weights $\beta_t\in (0,1)$ to estimate $C$.
    \\Probability distribution on $k\in \N$, the number of goals in the
    sparse rewards.
    \\Probability distribution on weights $w_i$ (default: $N(0,1)$),
    scaling factor $c_k$ (default: $1/\sqrt{k}$).
    \STATE \textbf{Output:}\\
    Trained features $\phi_1,\ldots,\phi_d$ with their covariance
    matrix $C$.
    \\Trained policies $\pi_z$.
    \WHILE{not done}
    \STATE Update covariance matrix $C$ via EMA:
    $C_{ij}\gets \beta_t
    C_{ij}+(1-\beta_t)\E_{s\sim \rho} \phi(s)\transp{\phi(s)}$
    Sample a value of $k$. Sample $k$ goal state-actions
    $(s^\star_i,a^\star_i)$ from the dataset distribution $\rho$. Sample
    weights $w_i$.
    \STATE Sample a state $s'\sim \rho$
    \STATE Compute
    \algindent
    $z(\phi)=\sum_i c_k w_i \bar C^{-1}
    \phi(s^\star_i)+c_k w_i \bar C^{-1}
    \left(\bar\phi(s')\transp{\bar\phi(s')}-\phi(s')\transp{\phi(s')}\right)\bar
    C^{-1}\bar \phi(s^\star_i)$
    \\where $\bar C$ and $\bar \phi$ are stop-grad versions of $C$ and
    $\phi$
    \STATE Update a $Q$-function $Q(s,a,z)$ at $z=z(\phi)$ with the Bellman loss
    \algindent
    $\ell(Q)=Q(s_t,a_t,z)^2-2\sum_i c_kw_iQ(s^\star_i,a^\star_i,z)-2\gamma
    Q(s_t,a_t,z)\bar Q(s_{t+1},a_{t+1},z)$
    where $(s_t,a_t,s_{t+1})$ is sampled from $\rho$, where $a_{t+1}$ is
    sampled from $\pi_z(s_{t+1})$, and where $\bar Q$ is a target version
    of $Q$.
    \STATE Update a policy $\pi_z(a|s)$ based on $Q(s,a,z)$, using any RL
    policy algorithm
    \STATE Update the occupation measure model $d(s,z)$ via one step of
    Algorithm~\ref{algo:occupation}
    \STATE Update $\phi$ with the loss
    \algindent $\loss(\phi)=-\sum_i c_k w_i\,d(s^\star_i,z(\phi))$
    \\where the gradients w.r.t.\ $\phi$ are backpropagated through $d$ and
    $z$.
    \ENDWHILE
    \STATE \textbf{Deployment:}
    \\Once the reward function $r$ is known:
    \STATE Estimate $\langle r,\phi_1\rangle_K$, \ldots $\langle
    r,\phi_d\rangle_K$
    \STATE Set $z=C^{-1}\langle r,\phi\rangle_K$
    \STATE Apply policy $\pi_z$
 \end{algorithmic}
 \end{algorithm}

\subsection{Mixing Priors}
\label{sec:mixing}

The zero-shot RL loss \eqref{eq:mainloss} is linear in the prior $\beta$.
Therefore, if two priors $\beta_1$ and $\beta_2$ are amenable to gradient
descent for this loss, one can deal with a mixture prior just by mixing
the losses for $\beta_1$ and $\beta_2$,
using a single set of features $\phi$, $Q$-functions, and
policies $\pi_z$.

In practice, this just means choosing at random, at each step, between
doing an optimization steps for one of the priors, e.g., alternating
between Algorithms~\ref{algo:main} and \ref{algo:sparse}.

Of course, this requires using consistent optimization methods for both
priors: the same optimizer, but also similar Bellman losses and policy
updates for $\beta_1$ and $\beta_2$. For instance, $\beta_1$ and
$\beta_2$ may both use the standard Bellman loss
$\left(Q(s_t,a_t,z)-r(s_t,z)-\gamma \bar Q(s_{t+1},a_{t+1},z)\right)^2$
where $r(s,z)$ is the mean reward knowing $z$ for a given prior. Then if
$\beta_1$ has posterior mean reward $r_1(s,z)$ knowing $z$ and likewise
for $\beta_2$, optimizing the $Q$-function alternatively between
$\beta_1$ and $\beta_2$ effectively optimizes for the mean posterior
reward of the mixture.

\section{Discussion}

\subsection{What Kind of Features are Learned? Skill Specialization and the Zero-Shot RL Loss}
\label{sec:whatfeatures}

The features
learned are influenced by the prior, and this is one reason why mixing
priors may be appealing.

For a pure goal-oriented prior, it is enough to learn a feature that
represents different goals by different values of $z$, so, it is enough
for $\phi$ to be injective (e.g., with $\dim \phi= \dim s$ and
$\phi=\Id$). On a discrete space, a one-dimensional $\phi$ may solve the
problem just by sending every state to a different value. Of course, this
will not work when mixed with other types of rewards.

For dense Gaussian priors, on the other hand, learning may produce
narrow features $\phi$, resulting in overspecialized skills. Indeed, 
conceptually, from Theorem~\ref{thm:main}, gradient descent of $\ell$ for $\pi_z$ and $\phi$ amounts
to:
\begin{itemize}
\item Learn $\pi_z$ to optimize reward $\transp{\phi}z$ for each $z$;
\item Learn $\phi$ by increasing $\transp{\phi}z$ at the states visited
by $\pi_z$.
\end{itemize}

The above is related to diversity methods \cite{eysenbach2018diversity}
and has a ``rich-get-richer'' dynamics: this is good for diversifying and
specializing, but might overspecialize. We illustrate this phenomenon
more precisely in the next paragraph.

\paragraph{Understanding overspecialization: Analysis with only one
feature, and influence of the prior.}
This is best understood on a ``bandit'' case (we can jump directly to any
state) and with only one feature.
In this case, a full analysis can be done, and the optimal
one-dimensional feature has only two non-zero values: a large positive
value at a state $s_1$ and a large negative value at a state $s_2$.
\footnote{Indeed, take
a finite state space $S=\{1,\ldots,n\}$ and assume that at any state,
there's an action directly leading to any other state: this makes the MDP
into a bandit problem.
Take $1$-dimensional $\phi$. Then from
Theorem~\ref{thm:main}, the gradient with respect to $\phi$ is
$d_{\pi_z}.z$ where $\pi_z$ is the policy to maximize reward $z.\phi$. If
$z>0$ then $\pi_z$ goes to the maximum of $\phi$, and
$d_{\pi_z}=(1-\gamma)U+\gamma \1_{\argmax \phi}$ where $U$ is the uniform
distribution. If $z<0$ then $d_{\pi_z}=(1-\gamma)U+\gamma \1_{\argmin
\phi}$. Since the distribution of $z$ is symmetric, on average the
gradient w.r.t.\ $\phi$ is proportional to $\1_{\argmax
\phi}-\1_{\argmin\phi}$. Gradient ascent on $\phi$ will converges to a
$\phi$ that has only two nonzero values, one positive and one negative.
(The cases where there are ties between the values of $\phi$ at several
states are numerically unstable.)
This applies to any Gaussian prior on rewards.
}

Is this specific to the ``bandit'' case? If the environment has full
reachability (the agent can reach any state and stay there), and if
the discount factor $\gamma$ is close to $1$, then the problem is
essentially a bandit problem. The transient dynamics before reaching a
target state will contribute $O(1-\gamma)$ to occupation measures
$d_\pi$, any the analysis done on the bandit case will hold up to
$O(1-\gamma)$.

Using a smoother prior on rewards (such as the Dirichlet prior, which
favors spatially smooth rewards) does not change this: this applies to
any Gaussian prior including the Dirichlet prior. The prior will
influence the location of the two states $s_1$ and $s_2$ at which the
feature is nonzero. \footnote{Intuitively, the feature $\phi$ only looks
at the reward at states $s_1$ and $s_2$ before choosing and applying a
policy.  With a Dirichlet prior, nearby states have correlated rewards,
so looking at the reward at $s_2$ does not bring much information if
$s_2$ is close to $s_1$: it brings more information to measure the reward at distant states
$s_1$ and $s_2$.
}

Yet such features \emph{are} optimal for the zero-shot RL loss with a
Gaussian prior. In an environment with full reachability and
$\gamma$ close to $1$, \emph{the optimal zero-shot behavior with one
feature consists in measuring the reward at two states and going to
whichever of those two states has the largest reward}. This applies to
any Gaussian prior on rewards, including priors whose covariance matrix
produces spatial smoothness on rewards.

So, if these features are considered undesirable, this
reflects a mismatch between the prior $\beta$ and the true distribution
of test tasks in the test loss \eqref{eq:testloss}. This pushes towards
mixing different types of priors, such as Gaussian and sparse reward priors.

Sparse reward priors such as goal-oriented (Dirac) rewards
correspond to smoother features such as $\phi=\Id$. This illustrates the
mathematical duality between $\phi$ and $r$ when estimating $z$ via
$\E[r.\phi]$: smoother priors on $r$ may lead to \emph{less} smooth
features $\phi$ (the dual of a space of smoother functions contains less
smooth functions). Intuitively, with sparse rewards $r$, the features
$\phi$ must be able to ``catch'' the location of the reward anywhere in
the space, and cannot be zero almost-everywhere.

\paragraph{Does the Bayesian viewpoint regularize the optimal features?} One might
have expected that the Bayesian flavor of the zero-shot RL objective
would result in regularized policies. But this is not the case: by
Proposition~\ref{prop:posteriormean}, every policy $\pi_z$ is a ``sharp''
policy, in the sense that it is optimal for some reward $r_z$.
Uncertainty on the reward does not induce noise on the policy: if the
maximum of $r_z$ is reachable, $\pi_z$ will go straight to it and stay
there. This contrasts with the effect of regularizations such as an
entropy regularization, which adds noise to the policy.

Yet this is the ``correct'' (optimal) answer given the zero-shot RL loss.

This overspecialization tendency has already been observed for diversity
methods: for instance, \cite{eysenbach2021information} also find that
skills learned must be optimal for some particular downstream task,
although they work from an information criterion and not from the
zero-shot RL loss. This seems to be an intrinsic property of this general
approach.

This illustrates the main assumption in the zero-shot RL framework: at test
time, the reward function is fully known and one can compute $z=\Phi(r)$.
This leaves no space for uncertainties on $r$, or any fine-tuning based
on further reward observations. The model estimates $z$, then applies a
policy that will maximize the mean posterior reward $r_z$, e.g., by going
to the maximum of $r_z$ and staying there if possible.

This optimal only if no uncertainty exists on $r$ and no fine-tuning of
the policies is possible.

\paragraph{Comparison with the Forward-Backward framework.} The results
in this text shows that forward-backward representations \cite{zeroshot,
allpolicies} have no reason
to be optimal: If the prior $\beta$ on tasks is known, then one shold
optimize the features for that prior.

However, the discussion above shows that the priors for which we can
compute optimal features may not necessarily reflect the kind of features
we expect to learn. Mixing different priors should mitigate that effect,
but to what extent is currently unclear.

Forward-backward representations aim at learning features that can
faithfully represent the long-term dynamics (successor measures) of many
policies. This is a different kind of implicit prior, closer in spirit to
a world model, and with no explicit distribution over downstream tasks.

\subsection{Future Directions}
\label{sec:future}

\paragraph{Avoiding overspecialized skills.} The zero-shot RL loss can
lead to very narrow optimal features with a Gaussian prior, as we have
seen. This is optimal for the loss \eqref{eq:mainloss}, but not what we
want in general, possibly reflecting a mismatch between a Gaussian prior
and ``interesting'' rewards.

One possible solution is to mix different priors.

Another possible solution is to account for variance over downstream
tasks in the zero-shot loss: we not only want the best expected
performance, but we don't want performance to be very bad for some tasks.
For the white noise prior, a full analysis is possible
(Appendix~\ref{sec:varianceregul}): incorporating variance is equivalent
to penalizing the $L^2$ norm of the occupation measures $d_\pi$ (this
will minimize spatial variance, thus ``spreading'' $d_\pi$). However,
it is not obvious how to exploit this algorithmically (since $d_\pi$ is
computed from $\pi$ and not the other way around, adding a penalty on
$d_\pi$ will just make the computation of $d_\pi$ wrong). Things are
actually simpler if we add a downstream task variance penalty to the FB
framework (Appendix~\ref{sec:varianceregul}).

Other solutions are to explicitly regularize the features (e.g., minimize
their spatial variance, or their Dirichlet norm to impose temporal
smoothness) or the policies (e.g., by entropy regularization). But since
the overspecialized features actually optimize the zero-shot RL loss (for
some priors), it is more principled to regularize the loss itself.

\paragraph{Nonlinear task representations.} In this text, we have covered
linear task representations, as these are the ones in the main zero-shot
frameworks available (successor features and forward-backward). However,
a linear task representation $r\mapsto z$ clearly limits the
expressivity of zero-shot RL.

One way to get nonlinear
reward representations, introduced in \cite{fb-aware}, is to iterate
linear reward representations in a hierarchical manner:
\begin{equation}
z_1=\E\, r(s)\phi_1(s),\qquad z_2=\E\,r(s) \phi_2(s,z_1),\qquad z_3=\ldots
\end{equation}
namely, $z_1$ provides a rough first reward representation, which can be
used to adjust features 
$\phi_2$ more precisely to the reward function. \cite{fb-aware} prove
that two such levels already provides full expressivity for the correspondence $r\mapsto z$.
This is amenable to a similar analysis as
the one performed in this text. The sparse reward case looks largely
unchanged, but the case of Gaussian priors is more complex: the
covariance matrix $C$ now depends on $z_1$, so it would have to be
represented via a learned model or estimated on a minibatch. We leave
this for future work.

Another way to bypass the linearity of the task representation would be
to kernelize the norm $\norm{r}_K$ used in the definition of Gaussian
reward priors.

\paragraph{Incorporating fine-tuning and reward uncertainty at test
time.} Finally, the analysis here relies the main hypothesis behind
the zero-shot RL framework: that at test time, the reward function is
instantly and exactly known. This is the case in some scenarios (eg,
goal-reaching, or letting a user specify a precise task), but not all. In
such situations, some fine-tuning of the policies will be necessary.
Which features provide the best initial guess for real-time fine-tuning is
out of the scope of this text. Zero-shot RL assumes the reward function is fully
specified at test time: if it is not, then meta-RL approaches
\cite{beck2023survey}
probably provide a better solution.

\section{Conclusions}

The zero-shot RL loss is the expected policy performance of a zero-shot
RL method on a distribution of downstream tasks. We have shown that this
loss is algorithmically tractable for a number of uninformative priors on
downward tasks, such as white noise, other Gaussian distributions
favoring spatial smoothness, and sparse reward priors such as
goal-reaching or random combinations of goals. We recover VISR as
a particular case for the white noise prior. We have also illustrated how
dense Gaussian reward priors can lead to very narrow optimal features,
which suggests that a mixture of different priors could work best.

\appendix

\section{Additional Proofs}
\label{sec:proofs}

\begin{dem}[ of Proposition~\ref{prop:posteriormean}]
This is because
$Q$-functions are linear in $r$ for a given policy. Intuitively, all
rewards represented by $z$ will share policy $\pi_z$, and so the
average return over rewards is the return of the average reward among
those represented by $z$. More precisely, by definition of $\beta_z$,
and by \eqref{eq:Vsucc},
the loss rewrites as
\begin{align}
\ell_\beta(\Phi,\pi)&=- \E_{z\sim \beta_z} \,\E_{r|\Phi(r)=z}\,\E_{s_0\sim \rho_0}\,
V^{\pi_z}_r(s_0)
\\&=- \frac{1}{1-\gamma}\E_{z\sim \beta_z}\, \E_{r|\Phi(r)=z}\,\E_{s\sim d_{\pi_z}}\,
 r(s)
\\&=- \frac{1}{1-\gamma}\E_{z\sim \beta_z}\, \E_{s\sim d_{\pi_z}}
\left(\E_{r|\Phi(r)=z}\, r(s)\right)
\\&=- \frac{1}{1-\gamma}\E_{z\sim \beta_z}\, \E_{s\sim d_{\pi_z}}
\, r_z(s)
\\&=- \E_{z\sim \beta_z}\, \E_{s_0\sim \rho_0}
\,V^{\pi_z}_{r_z}(s_0)
\end{align}
as needed.
\end{dem}

\begin{dem}[ of Propositions~\ref{prop:linearpostmean} and
\ref{prop:gaussianprior}]
Since Proposition~\ref{prop:linearpostmean} is a particular case of
Proposition~\ref{prop:gaussianprior}, we only prove the latter.

By definition, the reward $r$ is a centered
Gaussian vector with probability density $\exp(-\norm{r}^2_K/2)$.

The posterior mean reward $r_z$ is the expectation of $r$ knowing
\begin{equation}
z=C^{-1}\langle r,\phi\rangle_K
\end{equation}
where
$\langle
\cdot,\cdot\rangle_K$ is the dot product associated with the quadratic
form $\norm{\cdot}^2_K$, and
$C=\langle \phi,\phi\rangle_K$
is the $K$-covariance matrix of the features $\phi$,namely
$C_{ij}=\langle \phi_i,\phi_j\rangle_K$.

Without loss of generality, by the change of
variables $\phi\gets C^{-1/2}\phi$ (which yields $z\gets C^{1/2}z$), we
can assume that $C=\Id$, namely, the features $\phi$ are $K$-orthonormal. So we must compute
the mean of $r$ knowing $z=\langle r,\phi\rangle_K$.

Since $\phi$ is $k$-dimensional, this is a set of $k$ constraints
$\langle r,\phi_1\rangle_K=z_1, \ldots, \langle
r,\phi_d\rangle_K=z_d$.

These $k$ constraints define a codimension-$k$ affine hyperplane in the space of
reward functions. We have to compute the expectation of $r$ conditioned
to $r$ lying on this hyperplane.

For any Euclidean norm $\norm{\cdot}$, 
the restriction of a centered Gaussian distribution $\exp(-\norm{x}^2/2)$
to an affine subspace is again a Gaussian distribution, whose mean is
equal to the point of smallest norm in the subspace. (This can be proved
for instance by applying a rotation so the affine subspace aligns with
coordinate planes, at which point the result is immediate.)

Therefore, the posterior mean $r_z$ is the reward function that minimizes
$\norm{r_z}^2_K$ given the constraints $\langle
r,\phi_1\rangle_K=z_1, \ldots, \langle r,\phi_d\rangle_K=z_d$.
Since the $\phi_i$ are $K$-orthonormal, this is easily seen to be
$z_1\phi_1+\cdots+z_d \phi_d$. This proves the claim about the posterior
mean reward $r_z$.

For the claim about the distribution of $z$, assume again that the set of features
$\phi$ is $K$-orthonormal ($C=\Id$). Completing $\phi$ into a $K$-orthonormal
basis, the Gaussian prior $\exp(-\norm{r}^2_K)$ means that all components
of $r$ onto this basis are one-dimensional standard Gaussian variables.
So $z=\langle r,\phi\rangle_K$ is a $k$-dimensional standard Gaussian.
Undoing the change of variables with $C$, namely, $\phi \gets C^{1/2}\phi$ and
$z\gets C^{-1/2}z$, results in $z$ having covariance
$C^{-1}$.
\end{dem}

\begin{dem}[ of Theorem~\ref{thm:main}]
Theorem~\ref{thm:main} is a direct consequence of
Proposition~\ref{prop:gaussianprior},
Proposition~\ref{prop:posteriormean}, the definition of $\ell_\beta$, and
the expression \eqref{eq:Vsucc} of
$V$-functions using the occupation measures $d_\pi$.
\end{dem}

\paragraph{Derivation of Algorithm~\ref{algo:occupation}.}
The \emph{successor measure} \cite{successorstates} of a policy $\pi$ is
a measure over the state space $S$ depending on an initial state-action
pair $(s_0,a_0)$. It encodes the expected total time spent in any part
$X\subset S$, if starting at $(s_0,a_0)$ and following $\pi$. The formal
definition is
\begin{equation}
M^{\pi}(s_0,a_0,X)\deq \sum_{t\geq 0} \Pr(s_t\in X|s_0,a_0,\pi)
\end{equation}
for each $X\subset S$.

By definition, the occupation measure \eqref{eq:dpi} is the average of
the successor measure over the initial state and action:
\begin{equation}
\label{eq:dm}
d_\pi(X)=(1-\gamma) \E_{s_0\sim \rho_0, \, a_0\sim \pi(a_0|s_0)}
M^\pi(s_0,a_0,X).
\end{equation}

A parametric model of $M^\pi$ can be learned through various measure-valued Bellman
equations satisfied by $M$. For instance, TD learning for $M$ is
equivalent to the following.

Represent $M$ by its density with respect to the data distribution
$\rho$, namely
\begin{equation}
M^{\pi}(s_0,a_0,\d s)=m^\pi(s_0,a_0,s)\rho(\d s)
\end{equation}
where we want to learn $m^\pi(s_0,a_0,s)$.
This can be done using one of the methods from
\cite{successorstates}. For instance, $M^\pi$ satisfies a measure-valued
Bellman equation, which gives rise to a Bellman-style loss on $m^\pi$
with loss
\begin{multline}
\label{eq:lossm}
\loss_m\deq \E_{
(s_t,a_t,s_{t+1})\sim \rho,\,
a_{t+1}\sim \pi_z(s_{t+1}),\,
s'\sim \rho
}\\
\left[
\left(m^\pi(s_t,a_t,s')-\gamma
m^\pi(s_{t+1},a_{t+1},s')\right)^2-2m^\pi(s_t,a_t,s_t)
\right].
\end{multline}
This is the loss $\loss(M)$ used in
Algorithm~\ref{algo:occupation}, where an additional $z$ parameter
captures the dependency on $\pi_z$.

In turn, the relationship \eqref{eq:dm} between $d$ and $M$ can be used
to learn a parametric model of $d$ from a parametric model of $M$.
Let us parameterize  $d$ by its density with respect to $\rho$ as we did
for $M$, namely
\begin{equation}
d_{\pi_z}(\d s)=d(s,z)\rho(\d s)
\end{equation}
we find
\begin{equation}
d(s,z)=(1-\gamma) \E_{s_0\sim \rho_0, \, a_0\sim \pi(a_0|s_0)}
m(s_0,a_0,s,z)
\end{equation}
which provides the loss for $d$ in Algorithm~\ref{algo:occupation}.

\paragraph{Derivation of $\loss_C$.} This is essentially an application
of the log-trick $\partial_\theta \E_{z\sim p_\theta} f(z)=\E_{z\sim
p_\theta} [f(z)\partial_\theta \ln p_\theta(z)]$, as follows.

\begin{lem}
\label{lem:gradexpN}
Let $f\from \R^d\to \R$ be a bounded function and let $C$ be a $d\times
d$ matrix. Then the derivative with respect to $C$ of $\E_{z\sim
N(0,C^{-1})} f(z)$ satisfies
\begin{align}
\partial_C \E_{z\sim N(0,C^{-1})} f(z)=\tfrac12 \partial_C \E_{z\sim N(0,\bar C^{-1})}
\left[
f(z) \left(
-\transp{z}Cz+\Tr(\bar C^{-1}C)
\right)
\right]
\end{align}
taken at $\bar C=C$ (namely, $\bar C$ is a stop-grad version of $C$).
\end{lem}

Applying this to $C=\langle\phi,\phi\rangle_K$ yields
\begin{align}
\partial_\phi \E_{z\sim N(0,C^{-1})} f(z)&=
\tfrac12
\partial_\phi \E_{z\sim N(0,\bar C^{-1})}
\left[
f(z) 
\sum_{ij} \left((\bar C^{-1})_{ij}-z_iz_j\right)\langle
\phi_i,\phi_j\rangle_K
\right]
\end{align}
as needed for $\loss_C$.

\begin{dem}[ of Lemma~\ref{lem:gradexpN}]
For any smooth parametric probability distribution $p_\theta$ and any bounded
function $f$, one has the log-trick identity
\begin{equation}
\partial_\theta \E_{z\sim p_\theta} f(z)=\E_{z\sim p_\theta}[f(z)\partial_\theta \ln
p_\theta(z)].
\end{equation}
Here we have $\theta=C$ and
\begin{equation}
\ln p_\theta(z)=-\frac12 \transp{z}Cz+\frac12 \ln \det C+\mathrm{cst}
\end{equation}
and Jacobi's formula for the derivative of the determinant states that
\begin{equation}
\partial_C \det C=\partial_C \left(
(\det \bar C) \Tr(\bar C^{-1}C)\right)
\end{equation}
evaluated at $\bar C=C$, hence
\begin{equation}
\partial_C \ln \det C=\partial_C \Tr(\bar C^{-1}C)
\end{equation}
which implies the result.
\end{dem}

\begin{dem}[ of Proposition~\ref{prop:sparseloss}
]
Given a reward function $r$ and policy $\pi$, we have
\begin{align}
V^\pi_r&=\frac{1}{1-\gamma} \E_{s\sim d_\pi}r(s)
\\&=\frac{1}{1-\gamma} \E_{s\sim \rho} [d(s,\pi)r(s)]
\end{align}
where $d(s,\pi)$ denotes the density of $d_\pi$ with respect to $\rho$.

Applying this to a Dirac reward at $s^\star$, namely,
$r(s)=\1_{s=s^\star}/\rho(s^\star)$, yields
\begin{equation}
V^\pi_r=\frac{1}{1-\gamma} d(s^\star,\pi).
\end{equation}
The computation is the same when the reward is a sum of Dirac masses,
$r=c_k \sum_k w_i \delta_{s^\star_i}$
yielding
\begin{equation}
V^\pi_r=\frac{1}{1-\gamma} \sum_i c_k w_i d(s^\star_i,\pi).
\end{equation}

Proposition~\ref{prop:sparseloss} then follows from the definition of the
zero-shot loss $\ell_\beta$.
\end{dem}

\section{Penalizing Variance over the Reward $r$ is Equivalent to Spatial
Regularization for the White Noise Prior}
\label{sec:varianceregul}

The loss \eqref{eq:mainloss} maximizes the expected performance over the
reward $r$, but does not account for variance. This is one of the reason
we might get overspecialized skills that take ``risks'' such as making
a bet on the location of the best reward and going there.

Instead, let us consider a variance-penalized version of this loss,
\begin{equation}
\label{eq:mainlosspen}
\ell(\Phi,\pi)\deq - \E_{r\sim \beta}\, \E_{s_0\sim \rho_0}\,
V^{\pi_{\Phi(r)}}_r(s_0) + \lambda \Var_{r\sim \beta} (\E_{s_0\sim \rho_0}\,
V^{\pi_{\Phi(r)}}_r(s_0))
\end{equation}
where $\lambda\geq 0$ is the regularization parameter.

This is tractable, as follows. We only reproduce the main part of the
computation, the second moment term in the variance. With notation as in
Proposition~\ref{prop:posteriormean}, we have
\begin{align}
\E_{r\sim\beta} (\E_{s_0\sim \rho_0}\,
V^{\pi_{\Phi(r)}}_r(s_0))^2
&=
\E_{z\sim \beta_z} \E_{r| \Phi(r)=z} (\E_{s\sim \rho_0}\,
V^{\pi_z}_r(s_0))^2
\\&=\frac{1}{1-\gamma}
\E_{z\sim \beta_z} \E_{r| \Phi(r)=z} (
\E_{s\sim d_{\pi_z}} r(s)
)^2
\\&=\frac{1}{1-\gamma}
\E_{z\sim \beta_z} \E_{r| \Phi(r)=z} \langle d(s,z),r(s)\rangle_{L^2(\rho)}^2
\end{align}
where as in Section~\ref{sec:lossistractable}, $d$ is defined by
\begin{equation}
d_{\pi_z}(\d s)=d(s,z)\rho(\d s)
\end{equation}
namely, $d(s,z)$ is the density of the occupation measure of policy $\pi_z$
wrt the data distribution $\rho$.

For a white noise prior on $r$,
we have 
\begin{equation}
\E_r \langle d(\cdot,z),r\rangle_{L^2(\rho)}^2= \norm{d(\cdot,z)}^2_{L^2(\rho)}
\end{equation}
by the definition of white noise in general measure spaces.

But here we should not use $\E_r$ but $\E_{r|\Phi(r)=z}$, namely, we now
what the rewar features are. With a white noise prior and with linear
task representation, the distribution of $r$ knowing $\Phi(r)=z$ is the
$L^2(\rho)$-orthogonal projection of the white noise onto the orthogonal of the span of the
features. Denoting $\Pi^\bot_\phi$ this projector, we have
$\E_{r|\Phi(r)=z} \langle
d(\cdot,z),r\rangle_{L^2(\rho)}^2=\E_r \langle
d(\cdot,z),\Pi^\bot_\phi r\rangle_{L^2(\rho)}^2=\E_r \langle
\Pi^\bot_\phi d(\cdot,z),\Pi r\rangle_{L^2(\rho)}^2$. So, we have proved:

\begin{prop}
With linear features, penalizing the variance over $r$ of the expected
return is equivalent to penalizing the spatial variance (in
$L^2(\rho)$-norm) of the projection
onto the features of the occupation measure density $d(\cdot,z)$.
\end{prop}

Actually it might be safer just not to use the projection onto $\phi$:
it will overestimate variance, but results in simpler algorithms. Anyway,
the estimation of $z=\E[\phi.r]$ is itself subject to noise because we
use a finite number of samples, so even
knowing the empirical estimate of $z$, there is still variance in the
direction of the span of $\phi$.

Algorithmically, the applicability of this depends on the method. If the
occupation measure $d$ is just computed from $\pi_z$ which is computed
from $\phi$, then adding a penalty on $d$ will just throw off the
computation of $d$ without affecting the features $\phi$. But in the
VISR-like algorithm from Section~\ref{sec:lossistractable}, $\phi$ is in
turn computed from $d$ (the features $\transp{z}\phi$ are increased on
the part of the stte visited by $\pi_z$), so penalized the variance of $d$ is more or
less equivalent to penalizing the variance of $\phi$.

A similar penalty over the variance of the policy performance can be
incorporated in the FB framework. Things are a bit simpler because $B$ is
both the features and the successor measure $d$: we have
$d(s,z)=\E_{s_0\sim \rho_0,a_0\sim \pi_z(s_0)} \transp{F(s_0,a_0,z)}B(s)$, so
we might directly penalize the spatial variance of $d$, with loss
\begin{equation}
\E_{s\sim \rho} ((\E_{s_0,a_0} \transp{F(s_0,a_0,z)})B(s))^2
-(
\E_{s\sim \rho} \E_{s_0,a_0} \transp{F(s_0,a_0,z)}B(s)
)^2
\end{equation}

This may be a sensible and principled way to avoid degenerate
features in FB.

\bibliographystyle{alpha}
\bibliography{zeroshotloss}

\end{document}

%% file: header.en.tex
\usepackage{cmap}
\usepackage[utf8x]{inputenc}
\usepackage[T1]{fontenc}
\usepackage[english]{babel}
\usepackage{amssymb,amsmath}
\usepackage{amsthm}
\usepackage{bbm}
\usepackage{microtype}
\usepackage{lmodern}

{
\rmfamily
\DeclareFontShape{T1}{lmr}{m}{sc}{<->ssub*cmr/m/sc}{}
\DeclareFontShape{T1}{lmr}{b}{sc}{<->ssub*cmr/b/sc}{}
\DeclareFontShape{T1}{lmr}{bx}{sc}{<->ssub*cmr/bx/sc}{}
}

\input{header.tex}

\newenvironment{dem}[1][]{\begin{proof}[\thmheadercommand{Proof#1}]~\newline\ignorespaces}{\end{proof}}

{
\theoremstyle{yannthm}
\newtheorem{defi}{Definition}
\newtheorem*{defi*}{Definition}
\newtheorem{prop}[defi]{Proposition}
\newtheorem*{prop*}{Proposition}
\newtheorem{thm}[defi]{Theorem}
\newtheorem*{thm*}{Theorem}
\newtheorem{lem}[defi]{Lemma}
\newtheorem*{lem*}{Lemma}

\newtheorem*{cor*}{Corollary}

\newtheorem*{ex*}{Example}

\newtheorem*{subenonce}{}

\theoremstyle{yannthm2}

\newtheorem*{exo*}{Exercise}

\newtheorem{rem}[defi]{Remark}
\newtheorem*{rem*}{Remark}

\newtheorem*{subenonce2}{}

}

\newcommand{\transp}[1]{#1^{\!\top}\!}

%% file: header.tex
\newcommand{\thmheadercommand}[1]{\textbf{\scshape{}#1.\\*}}

\newtheoremstyle{yannthm}{\topsep}{\topsep}{\slshape}{}{\scshape\bfseries}{.}{.5em}{%
\thmname{#1}\thmnumber{ #2}\thmnote{#3}%
}
\newtheoremstyle{yannthm2}{\topsep}{\topsep}{}{}{\scshape\bfseries}{.}{.5em}{%
\thmname{#1}\thmnumber{ #2}\thmnote{#3}%
}

\def\d{\operatorname{d}\!{}}

\def\N{{\mathbb{N}}}

\def\R{{\mathbb{R}}}

\renewcommand{\geq}{\geqslant}
\renewcommand{\leq}{\leqslant}

\newcommand{\deq}{\mathrel{\mathop{:}}=}

\newcommand{\from}{\colon} 

\def\eps{\varepsilon}
\renewcommand{\epsilon}{\varepsilon}
\renewcommand{\phi}{\varphi}

\DeclareMathOperator{\Var}{Var}
\DeclareMathOperator{\Cov}{Cov}
\let\oldPr\Pr
\renewcommand{\Pr}{\oldPr\nolimits}
\newcommand{\E}{\mathbb{E}}

\DeclareMathOperator{\Tr}{Tr}

\DeclareMathOperator{\Id}{Id}

\DeclareMathOperator{\diag}{diag}

\newcommand{\abs}[1]{\left\lvert#1\right\rvert}
\newcommand{\norm}[1]{\left\lVert#1\right\rVert}

\newcommand{\1}{\mathbbm{1}}

\DeclareMathOperator*{\argmax}{arg\,max}
\DeclareMathOperator*{\argmin}{arg\,min}